\documentclass[review]{elsarticle}

\usepackage{hyperref}

\journal{Neurocomputing}

\usepackage{graphicx}
\usepackage{multirow}
\usepackage{times}
\usepackage{latexsym}
\usepackage{amsmath,amsthm}
\usepackage{amssymb}
\usepackage{amsfonts}
\usepackage[table]{xcolor}
\usepackage{colortbl}
\usepackage{algorithm}
\usepackage{mathrsfs}
\usepackage[noend]{algpseudocode}
\usepackage{booktabs} 
\usepackage{color}
\usepackage{natbib}
\usepackage{hyperref} 
\usepackage{url}
\usepackage{caption} 
\usepackage{algpseudocode}
\usepackage{pifont}
\usepackage{bm}
\usepackage{comment}
\usepackage{pdflscape}
\usepackage{bbding}
\usepackage{soul}
\usepackage{subfigure}

\def\1{{\bf{1}}}
\def\0{{\bf{0}}}

\begin{document}

\begin{frontmatter}


\title{ESIE-BERT: Enriching Sub-words Information Explicitly with BERT for Intent Classification and Slot Filling}

\author[swufe-address]{Yu Guo}
\author[swufe-x-address]{Zhilong Xie}
\author[swufe-x-address]{Xingyan Chen}
\author[swufe-sba-address]{Huangen Chen}
\author[swufe-address]{Leilei Wang}
\author[swufe-sba-address]{Huaming Du}
\author[swufe-sba-address]{Shaopeng Wei}
\author[swufe-x-address]{Yu Zhao}
\author[swufe-x-address]{Qing Li}
\author[swufe-ib-address]{Gang Wu\corref{cor1}}
\cortext[cor1]{Corresponding author:wugang@swufe.edu.cn}


\address[swufe-address]{Financial Intelligence and Financial Engineering Key Laboratory,\\ Southwestern University of Finance and Economics (SWUFE), Chengdu, 611130, China \\}
\address[swufe-x-address]{Financial Intelligence and Financial Engineering Key Laboratory,\\ Innovation Research Institute of Digital Economy and Interdisciplinary Sciences,\\ SWUFE, Chengdu, 611130, China \\}
\address[swufe-ib-address]{School of International Business, SWUFE, Chengdu, 611130, China \\}
\address[swufe-sba-address]{School of Business Administration, Faculty of Business Administration, SWUFE, China. \\}

\begin{abstract}
Natural language understanding (NLU) has two core tasks: intent classification and slot filling. 
The success of pre-training language models resulted in a significant breakthrough in the two tasks.
One of the promising solutions called BERT can jointly optimize the two tasks.
We note that BERT-based models convert each complex token into multiple sub-tokens by wordpiece algorithm, which generates a mismatch between the lengths of the tokens and the labels. 
This leads to BERT-based models do not do well in label prediction which limits model performance improvement.
Many existing models can be compatible with this issue but some hidden semantic information is discarded in the fine-tuning process.
We address the problem by introducing a novel joint method on top of BERT which explicitly models the multiple sub-tokens features after wordpiece tokenization, thereby contributing to the two tasks. 
Our method can well extract the contextual features from complex tokens by the proposed sub-words attention adapter (SAA), which preserves overall utterance information. 
Additionally, we propose an intent attention adapter (IAA) to obtain the full sentence features to aid users to predict intent.
Experimental results confirm that our proposed model is significantly improved on two public benchmark datasets.
In particular, the slot filling F1 score is improved from 96.1 to 98.2 (\textbf{2.1\% absolute}) on the Airline Travel Information Systems (ATIS) dataset.


\end{abstract}


\end{frontmatter}


\section{Introduction}
Since the twentieth century, the critical breakthrough of artificial intelligence and human language understanding technology has led to the successful deployment of various intelligent speech assistants, which are now ubiquitous in everyday life and used in various fields, such as assistants of fintech \cite{zhao2022Stock, zhao2022learning, tan2022finhgnn, huang2022asset}.
A fundamental technology of speech assistants is natural language understanding (NLU), the aim is to create a structured framework for user queries.
NLU generally includes two core tasks: {intent classification} and {slot filling} \cite{tur2011intent, tur2011spoken}.
The first task is usually a speaker intent detection issue, while the second task is a name entity recognition problem that predicts the sequence of slot labels.
For example, consider the query utterance \textit{"My phone is playing a lossless music."}, as shown in Table \ref{tab:example_for_id_sf}.
An entire utterance is associated with a specific intent, and each word has a different slot label in this utterance: the intent is \emph{PlayMusic}, "playing" is \emph{B-MusicPlay}, "lossless" is \emph{B-MusicType}, and "music" is \emph{I-MusicType}.

\begin{table}[htb]
    \centering
    \caption{An example sentence from the ATIS \cite{hemphill1990atis} dataset in IOB format (B indicates the start of the entity, I denotes its middle and end, and O refers to “outside the entity”). After wordpiece tokenization, “playing” is split into [’play’, ’\#\#ing’], and “lossless” is split into [’loss’, ’\#\#less’].
    }
    \resizebox{\linewidth}{!}{
    \begin{tabular}{c|c|c|c|c|c|c|c}
    \toprule   
        \textbf{Query} & \multicolumn{7}{c}{My phone is \textbf{playing} a \textbf{lossless} music} \\
        
    \midrule
        \textbf{Wordpiece tokenization} & \multicolumn{7}{c}{My phone is \textbf{play \#\#ing} a \textbf{loss \#\#less} music.} \\
    
    \midrule
        \multirow{2}{*}{\textbf{Slot Filling}}& My & phone& is& \textbf{play \#\#ing}& a& \textbf{loss \#\#less}& \textbf{music}\\ 
    \cmidrule{2-8}
         & O & O& O &\textbf{B-MusicPlay}& O& \textbf{B-MusicType}& \textbf{I-MusicType} \\
         
     \midrule
    \textbf{Intent Classification} & \multicolumn{7}{c}{PlayMusic} \\
    \bottomrule
        
    \end{tabular}
    }
    \label{tab:example_for_id_sf}
\end{table}

Many approaches implement natural language understanding separately. 
In the conventional approach, intention detection is treated as a text detection issue that is designed to predict the intent label. 
Each user-entered utterance has a specific intention.
Slot filling can be considered by name entity recognition task with each word in a sequence corresponding to a specific tag label.
The early intention classification and slot filling relied heavily on rules, e.g., \cite{ramanand2010wishful}. 
Machine learning methods were then explored for intent detection methods, particularly support vector machine (SVM) \cite{schuurmans2019intent}. 
As deep learning models have become increasingly diverse and mature, recurrent neural network techniques have enabled the NLU tasks for state-of-the-art (SOTA) performance, such as based long short-term memory (based-LSTM) and based-gated recurrent unit (based-GRU) models \cite{goo_slot-gated_2018, guo2014joint, hakkani2016multi}. 
Recently, a large number of pre-training language models have emerged with rich contextual information and excellent performance on NLU tasks.
One of the promising pre-trained solutions named BERT has facilitated SOTA method development \cite{devlin2018bert}.
Chen \textit{et al.} \cite{chen_bert_2019} devised a universal union method named joint BERT that encodes utterance semantics information by the BERT layer and then applies it to natural language understanding tasks.

We note that while the BERT-based methods are an effective strategy to joint the two NLU tasks, their problem is natural: they use subtoken-level word separation by the wordpiece algorithm. 
A complex token is tokenized into subtokens. For example, 'lossless' is tokenized into ['loss', '\#\#less'], as shown in Table \ref{tab:example_for_id_sf}. 
This leads to the lengths of tokens being not equal to the lengths of labels in the tag prediction, which inevitably lose some semantic information from the sentences and perform extremely poorly for the union model.
Therefore, it faces \textbf{two significant challenges}: 
\begin{itemize}
    \item \textbf{How to capture the entire contextual vector of complex tokens?}
    \item \textbf{How to predict labels without knowing the number of sub-tokens in advance?}
\end{itemize}

Some previous work has tried to address the two issues.
One of these is training a new BERT-style pretraining model.
WordBERT \cite{feng2022pretraining} is the promising one of them, which used word-level word separation and dispenses with the wordpiece algorithm.
WordBERT implemented SOTA performance for sequence labeling tasks given the corresponding prediction in the labeling. 
The others are improved approaches on top of BERT.
Joint BERT \cite{chen_bert_2019} fed a speaker's input tokens to a wordpiece tokenization layer, then used the output states corresponding to the start of the sub-tokens for the inputs to a slot filling decoder. 
AISE \cite{yang_aise_2021} introduced a novel position attend masks method to replace the softmax classifier, and also only used the output states corresponding to the start of the sub-tokens to predict labels.

\begin{figure}[htb]
    \centering
    \includegraphics[width=1\textwidth]{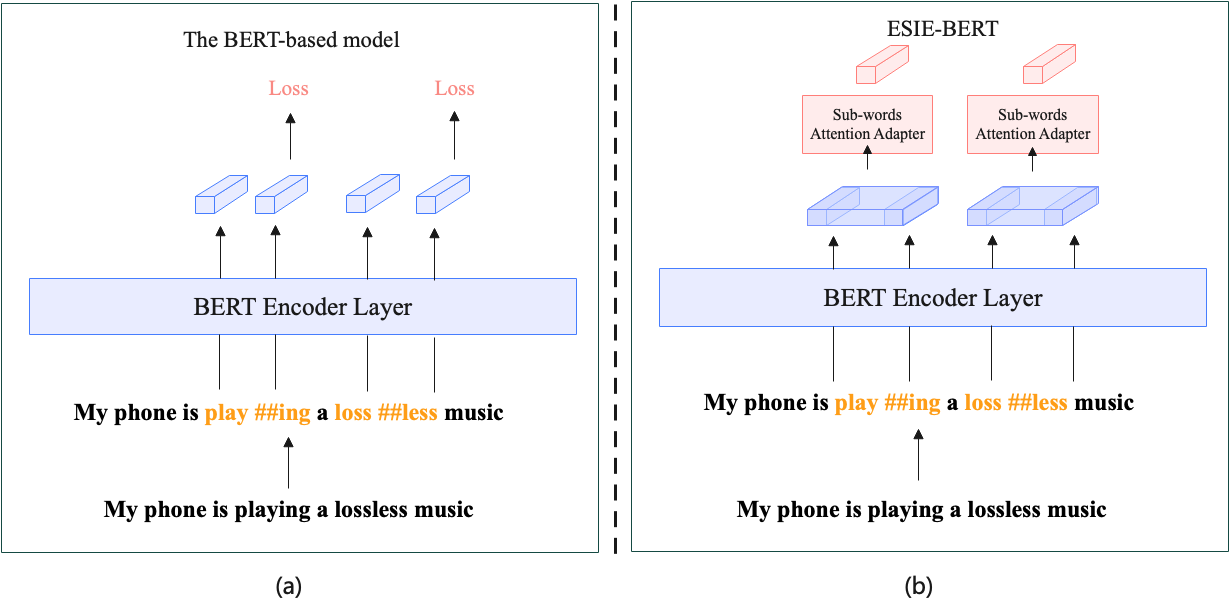}
    \caption{
    Example of a slot filling task that illustrates the difference between conventional BERT-based model \cite{chen_bert_2019, yang_aise_2021} and ESIS-BERT. In the BERT-based model,  some hidden information corresponding to subtokens is lost, whereas ESIS-BERT uses the SAA to take advantage of all subtokens information.}
    \label{fig:AIST-compare}
\end{figure}

However, these approaches lose some hidden states of information corresponding to sub-tokens and fail to take advantage of the complete utterance features. 
To address the problem, we introduce a novel method for joint natural language understanding tasks that use a \textbf{Sub-word Attention Adapter (SAA)} to explicitly model multiple sub-token features and an \textbf{Intent Attention Adapter (IAA)} to obtain the overall information of an utterance. 
Unlike the conventional BERT-based model, our model can capture associations between sub-tokens, thereby providing more token features for the NLU tasks.
Figure \ref{fig:AIST-compare} illustrates the difference between conventional BERT-based models and our proposed model using a slot filling example. 
As shown in Figure \ref{fig:my_label}, the proposed method uses a BERT layer to encode shared hidden features for two natural language understanding tasks.
Then, our method uses the SAA to capture each subword’s features and thereby obtains all subtokens information after wordpiece tokenization algorithms. 
In the SAA, a self-attention layer can be applied to extract features from each complex subtoken, as illustrated in Figure \ref{fig:my_label} (b). 
For intention classification task, our method uses IAA to obtain intent features from overall utterances.
The utterance intent features used in IAA are a weighted sum of the hidden states of single tokens and the contextual features of complex tokens.
For slot filling task, predictions for simple tokens consist of intent features and hidden states, and predictions for complex tokens consist of intent features and contextual features.
Finally, a softmax layer is used as the intent detection decoder and the slot filling decoder. 
Since the proposed model \textbf{E}nrich \textbf{S}ub-words \textbf{I}nformation \textbf{E}xplicitly and build on top of \textbf{BERT}, we name it \textbf{ESIE-BERT}. 

Experimental results show that the performance of ESIS-BERT is clearly superior to that of all the baseline methods on two benchmark datasets (ATIS \cite{hemphill1990atis} and SNIPS \cite{coucke2018snips}).
In particular, the slot filling performance of ESIS-BERT is significantly better, which is probably because ESIS-BERT can learn more features to meet the need to retain complete sentences.
In summary, our work makes three main contributions, as follows:

\begin{itemize}
    \item To the best of our knowledge, we are the first to analyze the wordpiece tokenization issue on top of BERT.
    We propose ESIS-BERT, a novel method for joint intent classification and slot filling that uses an SAA to explicitly model multiple sub-token features and an IAA to obtain an utterance overall feature. 
    
    \item We analyze the effectiveness of the SAA and IAA and conduct extensive experiments to demonstrate the ESIS-BERT. 
    The visualization of the learning weight scores demonstrates the interpretability of the model.

    \item The experiments show our ESIS-BERT performs significantly better than conventional SOTA models in intent classification accuracy, slot filling F1, and sentence-level accuracy on the ATIS and SNIPS datasets.
\end{itemize}

The rest of this paper is organized as follows. 
In section 2, many related works are presented to show how the proposed work is unique. 
In section 3, the architecture of ESIS-BERT is introduced. 
In section 4, we discuss the experiment datasets, training details, baselines, and evaluation results. 
In section 5, we analyze the effect of SAA and IAA for the proposed method, the ablation analysis, different epochs, a case study, and the visualization of SAA mechanism.
Section 6 concludes with conclusions from this proposed method.

\section{Related work}
In this section, we introduce many previous works on wordpiece problems in pre-trained language models.
We then introduce many NLU task methods, including intent classification, slot filling, and joint methods.

\subsection{Pre-training Without Wordpiece}
Pretraining language models can make it easy to model downstream tasks in NLP.
Traditional BERT-based models use subword-level word separation (e.g. the wordpiece algorithm).
Recently, Feng \textit{et al.} \cite{feng2022pretraining} proposed a solution called WordBERT, which dispenses with the wordpiece algorithm, as opposed to the traditional BERT-based model.
The WordBERT used word-level word separation and trained a new BERT-style pretraining model.
WordBERT exceeded the standard BERT for all NLP-related tasks, especially for sequence labeling tasks.

However, our approach differs from the previous works in that we are not training a BERT-style pretraining model, which would require significant GPU resources.
Our method model subword-level and overall sentence feature on top of BERT by the sub-words attention adapter and the intent attention adapter for the NLU tasks.

\subsection{Intention Classification}
Intention classification can be formulated as an utterance detection issue \cite{abdalla2022integration,cui2022self,wu2017efficient}. 
Many machine learning methods can be applied for text classification, particularly support vector machine (SVM) \cite{schuurmans2019intent}. 
As a text classification task, its performance is dependent on a complete utterance.
Conventional machine learning methods struggle to extract utterance features, limiting the performance of utterance detection.
Intention detection has been widely explored using deep neural network methods. 
Convolutional neural networks (CNN) have been introduced to classify the intent of a speaker utter.
Kim \cite{kim_convolutional_2014} reported a simple CNN that achieved excellent performance for text classification, which learned many specific task features through more fine-tunings.
Zhang \textit{et al.} \cite{zhang_character-level_2015} empirically explored a special CNN method for utterance detection.
In addition, RNNs and LSTMs have been used extensively in intention detection tasks.
Ravuri and Stolcke \cite{ravuri_recurrent_nodate} proposed RNN and LSTM models for utterance classification.
Many attention-based models have been used for intention classification \cite{zhao_attention-based_2016,yang_hierarchical_2016,zhang2023strengthened}.

However, we introduce a joint method for the two NLU tasks that are different from those that have been used in previous studies, as it applies the sub-words attention adapter and intent attention adapter to model overall sentence features and sub-word features of complex words.
Also, the model shares the information of its two tasks, such that its performance of each task is better than that obtained from the individual execution of the tasks.

\subsection{Slot Filling}
Slot filling can be formulated as a sequence labeling task. 
Many conventional methods use probabilistic methods such as HMM, MEMM \cite{mccallum2000maximum}, and CRF \cite{raymond2007generative} to solve the task. 
After, many solutions for sequence labeling tasks are deep artificial neural network models, in which the performance has surpassed that of methods based on conventional machine learning.
The CNN \cite{vu_sequential_2016} model was proposed to solve the sequence labeling task in spoken language understanding.
In addition, many based-RNN models have become popular approaches for sequence labeling, such as LSTM and RNN.
Yao \textit{et al.} \cite{yao_spoken_2014} applied LSTM to a word labeling task.
Then, Peng \textit{at al.} \cite{peng2015recurrent} used an RNN model for slot filling, which aids in memorization capability via many external memories.
Kurata \textit{at al.} \cite{kurata_leveraging_2016} were the first researcher to explicitly model label dependencies, which enhance LSTM-based word labeling. 
Zhao and Feng \cite{zhao_improving_2018} introduced the seq2seq model for sequential labeling.
He \textit{et al.} \cite{he2021context} used a knowledge integration mechanism on top of BERT for slot tagging.

The difference between these methods and our method is that the ESIE-BERT focuses more on each word in the overall utterance.
We use the sub-words attention adapter to learn features between sub-tokens for complex words.
The advantage of this is that each word (sample word and complex word) is classified using all its features.

\subsection{Joint Model}
In NLU tasks, intention and slot are highly correlated, so the two tasks can best be modeled as being related.
Many methods implicitly learn feature information in the intention detection and slot labeling issues by sharing parameters.
And then, other methods explicitly feed intent features to the slot tagging decoder, and other methods aim to extract the inter-relation for intention and slot.
Liu and Lane \cite{liu_attention-based_2016} introduced a seq2seq method that contains the attention mechanism. The model allows a multilayer perceptron machine to capture the associate with slot and intent. 
Goo \textit{at al.} \cite{goo_slot-gated_2018} devised a novel joint method by slot-gated model. The model explores the correlation between slot tagging and intention detection, which feeds intention features into slot tagging by the slot-gated method.
Zhang \textit{at al.} \cite{zhang_graph_2020} adopted a new method named Graph LSTM to convert utterance into a graph, in which the classical message-passing mechanism in graph neural networks is used to pass contextual information to the prediction of intents and locations.
Many models based Transformer models were proposed for the two NLU tasks. 
Qin \textit{at al.} \cite{qin_co-interactive_2021} developed a novel Co-Interactive Transformer method that considers the crossover influence of intention classification and slot tagging.
Wang \textit{at al.} \cite{wang2021encoding} devised a new Transformer encoder-based method. 
The method uses syntactical knowledge encoded for intention classification and slot filling.
Many graph-based methods \cite{tang2020end} were introduced for the two NLU tasks. 
Ding \textit{at al.} \cite{ding2021focus} constructed a Dynamic Graph Model (DGM) for intention classification and slot tagging. 
The model leverages an interactive graph to capture the inner relationship between intention with the slot.

In recent, many excellent pre-training methods have been used for the NLU tasks, such as BERT \cite{devlin2018bert}, ELMo \cite{DBLP:journals/corr/abs-1802-05365} and GPT \cite{radford2018improving}, by using an enormous amount of unannotated text to address data sparsity challenge. 
Chen \textit{at al.} \cite{chen_bert_2019} first used the BERT model to explore joint slot filling and intents classification.
Qin \textit{at al.} \cite{qin_stack-propagation_2019} used a new method named Stack-Propagation for NLU tasks to feed intention features into the slot decoder layer.
Also, the author of Stack-Propagation
 \cite{qin_stack-propagation_2019} used the BERT layer in the method, which further boosts the result of the NLU task.

However, most of them still lack solutions for managing sub-tokens after wordpiece tokenization. 
Our proposed model is the first to address wordpiece tokenization issue on top of BERT-based models.  
The sub-words attention adapter can extract contextual features, which preserves the integrity of a given sentence.
The intent attention adapter can capture an utterance feature, which aids the performance of intent.

\section{Proposed Model}
\begin{figure}[htb]
    \centering
    \includegraphics[width=1\textwidth]{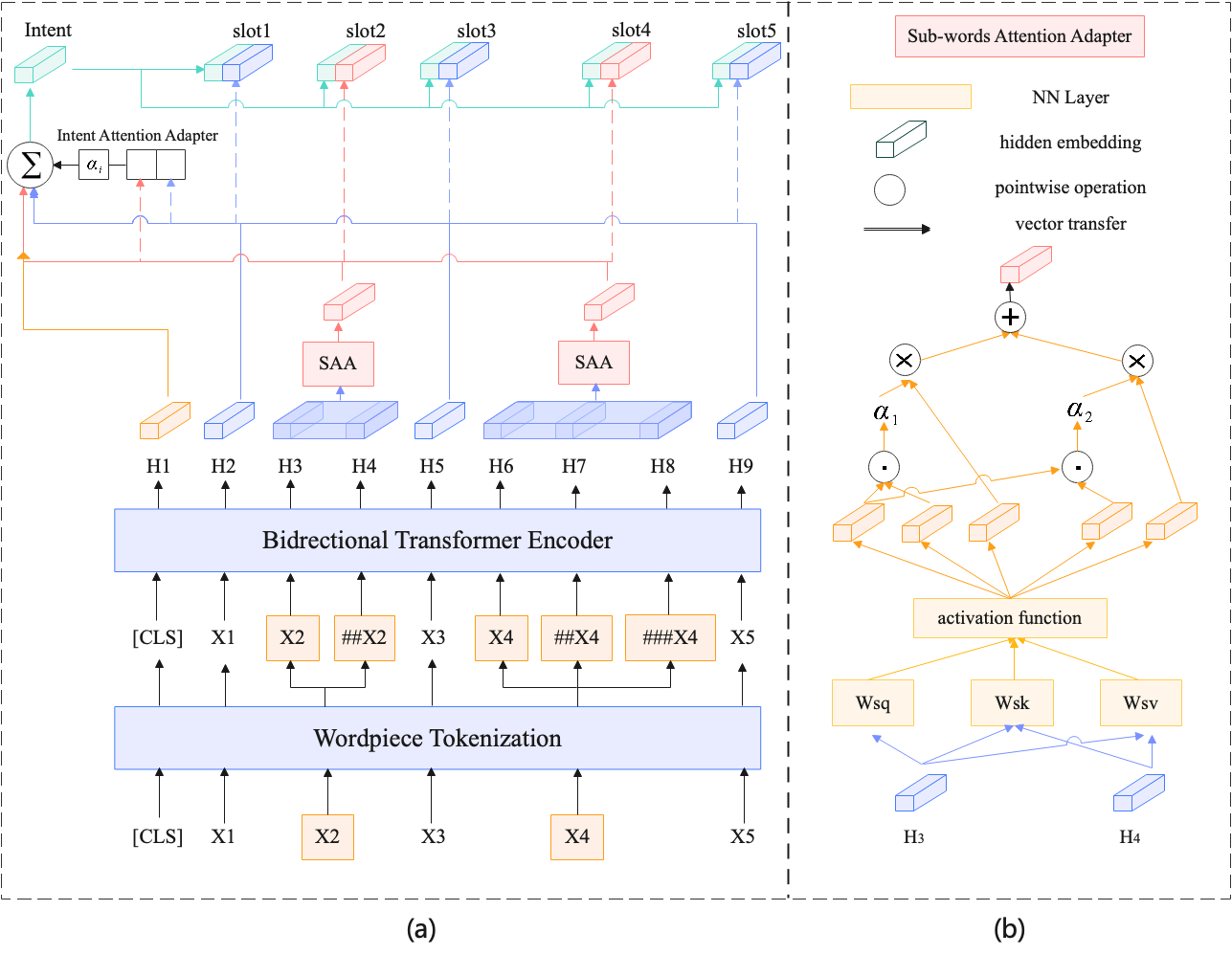}
    \caption{
    \textbf{Illustration of the ESIE-BERT for natural language understanding tasks.} 
    The main architecture, illustrated in (a), consists of a BERT layer, a sub-words attention adapter, an intent attention adapter, and two decoders (intent decoder and slot decoder). 
    (b) is the sub-words attention adapter that uses a self-attention layer to extract sentence semantic information for subwords. 
    \label{fig:my_label}
    }
\end{figure}

In this section, we introduce the ESIE-BERT for natural language understanding tasks. 
The ESIE-BERT is a multi-task joint training model and improves the base-BERT method in three folds: 
(1) we use sub-words attention adapter (SAA) to merge sub-words features for align tokens and tags (Section \ref{section-SAA}), 
(2) for intent classification, we leverage the intent attention adapter (IAA) to extract features from the hidden states for all tokens and the contextual features for all sub-words (Section \ref{section-IAA}),
(3) we feed the intention feature, BERT’s hidden states, and sub-words features into a slot filling decoder, and the ESIE-BERT method is jointly optimized for both intention detection and slot tagging (Section \ref{section-JL}).
In this paper, we define speaker utterance as model inputs $X = (x_1 , ..., x_T )$,  slot labels as $y^S = (y_1^S , . . . , y_T^S)$, intent label as $y^I$, and the hidden states of BERT layer as $H = (h_1 , ... , h_T )$, the overall architecture is illustrated in Figure \ref{fig:my_label}.

\subsection{Sub-words Attention Adapter}
\label{section-SAA}
This part describes the sub-words attention adapter (SAA) illustrated in Figure \ref{fig:my_label} (b). 
The ESIE-BERT uses the BERT encoder to encode a sentence in order to capture shared information common to the two NLU tasks. 
The BERT model makes use of the wordpiece algorithm resulting in a complex word that can be broken into sub-words.
Figure \ref{fig:AIST-compare} shows that ‘playing’ is mapping to ['play', '\#\#ing'] based on the wordpiece algorithm. 
This results in the length of tokens not matching the length of tags. 
Therefore, we propose the SAA to solve the alignment problem of tokens and tags.

The SAA uses a self-attention mechanism to construct complex word semantic features in multiple subwords, which ensures that the word information is complete. 
This attention mechanism uses all of the semantics information of the complex words, thereby retaining complete sentence semantic information to improve NLU task performance. 
For each complex word, we obtain the hidden states corresponding to its subwords by using the BERT encoder layer. 
Subsequently, for the hidden states corresponding to each subword, we use the learned attention weights to compute the subword contextual vector $s_i$:

\begin{equation}
    s_{i} = \sum_{i=1}^{T}\alpha_{i}v_{i} \ ,
\end{equation}
where $v_{i}$ is the value transform vector corresponding to each subword, the $\alpha_{i}$ is the subwords of complex word information score and computed as below.

\begin{equation}
\alpha _{i} = \frac{exp(e_{1,j})}{\sum_{j=1}^{T}exp(e_{1,j})} \ ,
\end{equation}
\begin{equation}
e_{1,j} = q_{1} \cdot k_{i} \ ,
\end{equation}
where $q_{1}$ is the query transform vector of the first subword of each complex word, and $k_{i}$ is the key transform vector corresponding to each subword. 

\begin{equation}
q_{i} = \sigma (W_{sq}h_{i}) \ ,
\end{equation}
\begin{equation}
k_{i} = \sigma (W_{sk}h_{i}) \ ,
\end{equation}
\begin{equation}
v_{i} = \sigma (W_{sv}h_{i}) \ ,
\end{equation}
where $\sigma$ is the activation function, $h_{i}$ is the hidden states corresponding to each subword and $W_{sq}, W_{sk}, W_{sv}$ are the weight matrix. 
In the following, we exploit the features of contextual subwords obtained by the SAA for slot prediction and intent recognition tasks.

\subsection{Intent Attention Adapter}
\label{section-IAA}
Given the diversity of human languages, different tasks typically require different matching models.
This leads to a limited generalization of deep learning technology.
With the emergence of pre-training technology, this problem has been successfully addressed, and only simple fine-tuning is required to achieve significant performance on downstream tasks.
These pre-training language models are heavily pre-trained, and each word has a strong contextually relevant utterance representation.  
One of the promising pre-training solutions is Encoder Representations from Transformers (BERT) \cite{devlin2018bert}, which is consisted of the standard Transformer architecture.
BERT inserts a special character \textit{[CLS]} at the start of each input utterance, which can typically represent the features of the full utterance. 
Therefore, for utterance classification tasks, the BERT model usually achieves meaningful performance by the hidden state of \textit{[CLS]}.
However, this approach has been the subject of much debate in recent literature \cite{lample2019cross,joshi2020spanbert,yang2019xlnet,lan2019albert}.
Even debating whether it should be removed\cite{hemphill1990atis}.

In the NLU tasks, using the hidden state corresponding to each word that can retain the global information of the sentence is better than using only the hidden states of the token [CLS].
Therefore, we use an intent attention adapter (IAA) to extract the intention features from the tokens of the sentence.
Namely, the global intent feature $h_{intent}$ is calculated as follows:

\begin{equation}
W(H)=
\left\{
\begin{array}{l}
h_1 \cdot \sigma(W_{int}h_i),\ i\ is\ not\ complex\ word,\\
h_1 \cdot \sigma(W_{int}s_i),\ i\ is\ complex\ word,\\
\end{array}
\right.
\end{equation}
where $W(H)$ is a weight function of each word, $\sigma$ is the activation function, $W_{int}$ is a weight matrix, $h_1$ is the hidden state of the [CLS], $h_i$ is the hidden state of the tokens, and $s_i$ is the sub-word contextual vector. 

\begin{equation}
\alpha _{intent} = softmax(\frac{W(H)}{\sqrt{d_h}}) \ ,
\end{equation}
where $d_h$ is the hidden size of the BERT.

\begin{equation}
h_{intent} = \sum_{i=1}^{T}\alpha _{intent}^ih_i \ ,
\end{equation}
Then, we use $h_{intent}$ directly to predict the user intention and insert it into the slot decoder to improve the performance of slot tagging task.

\subsection{Joint Learning}
\label{section-JL}

\noindent \textbf{Intent Classification} \
For the intent detection task, the ESIS-BERT feeds the global intent feature $h_{intent}$ from the IAA into the intention decoder, and the intent $y^{I}$ is predicted as:

\begin{equation}
    y^{I} = softmax(W^{i}(h_{intent} + h_1) + b^{i})\ ,
\end{equation}
where $W^i$ is a weight matrix, $h_1$ is the hidden state of the [CLS], and $b^i$ is a bias.

\noindent \textbf{Slot Filling} \
For the slot filling task, the ESIS-BERT uses the global intent feature $h_{intent}$ from the IAA and the sub-word contextual vector $s_{i}$ from the SAA to predict the tag label sequence $y^S$, the slot is predicted as:

\begin{equation}
y_{i}^{S} =
\left\{
\begin{array}{l}
softmax(W^{S}(h_{intent}+h_{i}) + b^{S}),\ i\ is\ not\ complex\ word,\\
softmax(W^{S}(h_{intent}+s_{i}) + b^{S}),\ i\ is\ complex\ word,\\
\end{array}
\right.
\end{equation}
where $y_{i}^{S}$ corresponds to the slot label for each word, $W^{S}$ is the weight matrix, and $s_{i}$ is the slot context vector of each word after the SAA.

\noindent \textbf{Optimization} \
Obviously, the ESIS-BERT can easily be extended to a joint model for the two core NLU tasks. 
Here, the ESIS-BERT employs a joint optimization learning process, we design a loss function $\mathcal{L}_{joint}$ as follows:

\begin{equation}
\mathcal{L}_{joint} = \beta * \mathcal{L}_i + (1 - \beta)\mathcal{L}_s
\end{equation}
where $\beta$ is a hyper-parameter by adjusting the interaction of the two tasks on the training, $\mathcal{L}_i$ and $\mathcal{L}_s$ apply cross-entropy loss over a softmax activation.
The objective of the ESIS-BERT is to minimize the score of $\mathcal{L}_{joint}$.

\section{Experiments}
\subsection{Experimental Datasets}
Our work adopts famous two NLU datasets: the ATIS \cite{hemphill1990atis} and SNIPS \cite{coucke2018snips} datasets. 
The ATIS dataset contains only flight information, while the SNIPS dataset is more complex, as it contains information on weather, restaurants, and music.
Both datasets are English-language datasets.
Table \ref{mytabel:dataset} shows the segmentation and labeling details for the ATIS and SNIPS.

\noindent \textbf{ATIS}: The ATIS dataset includes users' audios of flights and orders, consisting of 4,478, 893, and 500 sentences for the training set, validation set, and testing set, respectively. 
ATIS training data include 120 slot labels and 21 intent types, and 1,597 subwords.

\noindent \textbf{SNIPS}: The SNIPS dataset is collected through SNIPS Smart Audio and comprises 13,084, 700, and 700 utterances for the training set, validation set, and testing set, respectively. This dataset includes 72 slot labels and 7 intent types, and 10,511 subwords.

\begin{table}[H]
\centering
\caption{\textbf{Datasets segmentation and labeling details for ATIS and SNIPS datasets.}The first group is the background information of the dataset, the second group is the splitting of the dataset and the third group is the number of labels.}
\begin{tabular}{p{4cm} ||c || c}
\toprule
\textbf{Dataset} & \textbf{ATIS} & \textbf{Snips}  \\
\midrule
Vocabulary Size & 11241 & 722  \\
Average Sentence Length & 9.05 & 11.28  \\
Number of Subwords & 1597 & 10511  \\
\midrule
\#Training Set Size & 4478 & 13084  \\
\#Validation Set Size & 500 & 700  \\
\#Testing Set Size & 893 & 700  \\
\midrule
\#Slot labels & 120 & 72  \\
\#Intents & 21 & 7  \\
\bottomrule
\end{tabular}

\label{mytabel:dataset}
\end{table}

\subsection{Training Details}
Table \ref{tab:hpyerparameters} shows the experimental hyperparameters setting information. 
The ESIS-BERT uses English uncased Bert-Base model \cite{devlin2018bert} as the pre-training layer, which has 12 layers, 768 hidden states, and 12 heads. 
The attention mechanism has 768 hidden states, and the batch size is 256. 
The optimizer for model training uses AdamW \cite{kingma2014adam}, with a learning rate of 5e-5.
The dropout probability is 0.1. 
The different epochs are set as [10, 30, 40, 50, 60, 80]. 
The joint optimization has a $\beta$ value of 0.7 in training.
For fine-tuning process, all of the hyper-parameters are tuned on the validation dataset. 
All of the experiments are conducted on GeForce RTX 3090 GPU.

\begin{table}[htb]
    \centering
    \caption{Hyper-parameters of experiments.}
    \begin{tabular}{l||l||c}
    \toprule
        Layer & Hyper-parameter & Size \\
        \midrule\midrule
        \multirow{3}{*}{BERT Embedding} & Dimension & 768 \\ 
         & Layers & 12 \\ 
         & Heads & 12 \\ \midrule
        Attention & Hidden states & 768 \\ \midrule
        Dropout & Dropout rate & 0.1 \\ \midrule
         \multicolumn{2}{l||}{Batch size}  & 256 \\ 
         \multicolumn{2}{l||}{Number of epochs} & 30 \\ 
         \multicolumn{2}{l||}{Learning rate} & 5e-5 \\ 
         \multicolumn{2}{l||}{Training ($\beta$)} & 0.7 \\ 
    \bottomrule
    \end{tabular}
    \label{tab:hpyerparameters}
\end{table}

\subsection{Baselines}
We compare the ESIS-BERT with three types of baseline models:
 an implicit joint model \cite{liu_attention-based_2016}; 
 explicit joint models \cite{goo_slot-gated_2018, zhang_graph_2020, qin_co-interactive_2021, wang2021encoding, ding2021focus}; 
 and the pre-training models \cite{chen_bert_2019, qin_stack-propagation_2019}.
 All results are obtained using the same datasets.

\noindent \textbf{\textit{The implicit joint model:}}

\begin{itemize}
    \item \textbf{Attention BiRNN.} Attention BiRNN \cite{liu_attention-based_2016}  developed a seq2seq method that contains the attention mechanism. 
    The model allows a multilayer perceptron machine to capture the associate with slot and intent. 
\end{itemize}

\noindent \textbf{\textit{The explicit joint model:}}

\begin{itemize}
    \item \textbf{Slot-Gated Atten.} Slot-Gated Atten \cite{goo_slot-gated_2018} devised a new joint model by slot-gated mechanism, exploring the inner relationships between slot tagging with intention detection, which feeds intention features into slot tagging by the slot-gated method. 
    
    \item \textbf{Graph LSTM.} Graph LSTM \cite{zhang_graph_2020} adopts a novel method joint named Graph LSTM to convert utterance into a graph, in which the classical message-passing mechanism in graph neural networks is used to pass contextual information to the prediction of intents and locations.
    
    \item \textbf{Co-Interactive transformer.} 
    Co-Interactive transformer \cite{qin_co-interactive_2021} used a Co-Interactive Transformer method that considers the crossover impact of intention classification and slot tagging.
    
    \item \textbf{SyntacticTF.} SyntacticTF \cite{wang2021encoding} utilized a new encoder-decoder method based on Transformer.
    The method encodes information of syntax into the intention and slot decoder.
    
    \item \textbf{DGM.} 
    DGM \cite{ding2021focus} developed a DGM for intention classification and slot tagging.
    The model leverages an interactive graph to capture the relationship between intent and slot.

\end{itemize}

\noindent \textbf{\textit{The pre-training models:}}

\begin{itemize}
    \item \textbf{Joint BERT.} Joint BERT \cite{chen_bert_2019} first uses the BERT model to explore joint slot filling and intents classification. 
    The BERT performs well in downstream tasks as it has been pre-trained for a large number of tasks.
    
    \item \textbf{Stack-Propgation + BERT.} Stack-Propgation \cite{qin_stack-propagation_2019} introduced a Stack-Propagation method to feed intention features into slot decoder, in which the effect is further enhanced by the addition of a BERT layer.

\end{itemize}

\subsection{Experimental results}
Based on previous studies, we choose three metrics to evaluate the performance of two NLU tasks, consisting of intent classification accuracy, slot filling F1, and sentence-level accuracy, which is the proportion of sentences in the whole dataset in which slot and intent are both correctly predicted.
We choose the best parameters of the model on the validation set and then evaluate the final performances of the model on the test set.
Table \ref{tab:ex-result} compares the performances of the ESIS-BERT with the performances of the previous SOTA models.

In this table, the first method is the baseline of the implicit joint model, the second group of models is many explicit joint models, and the third group of models is the pre-training model.
These methods represent the SOTA joint intention detection and slot tagging models.
The final row is our proposed ESIS-BERT, comparing all of the baselines for the slot filling task, it clearly gets a significant improvement on both datasets.
Therefore, our proposed ESIS-BERT has become the most up-to-date SOTA model.

With the ATIS dataset, the ESIE-BERT achieves slot tagging F1 of 98.2\% (from 96.1\%), and sentence-level accuracy of 90.1\% (from 88.6\%). 
In the SNIPS dataset, the ESIE-BERT gets an intent accuracy of 99.1\% (from 99.0\%), and slot filling F1 of 97.8\% (from 97.0\%). 
This set of results demonstrates the effectiveness of the ESIS-BERT for the two NLU tasks.
This result shows that slot filling performance is significantly improved. 
We have summarised this improvement for two main reasons:

\begin{itemize}
\item 
The ESIE-BERT model uses the SAA to extract complex contextual features from sub-tokens, with leverages of the IAA to enrich the meaning of the sentence.
\item 
The prediction of slot tags relies on a clear correspondence between the token and the label, which the SSA can explicitly model the relationship.
\end{itemize}

\begin{table}[htb]
    \centering
    \caption{
    \textbf{Natural language understanding results on the SNIPS and ATIS datasets.} 
    The metrics are intent classification accuracy, slot filling F1, and sentence-level accuracy. 
    We adopt the main results of baselines reported by SyntacticTF(Independent) \cite{wang2021encoding}, DGM \cite{ding2021focus}, and the NLU survey \cite{qin_survey_2021}.
    }
    \resizebox{\textwidth}{!}{
    \begin{tabular}{l || ccc || ccc}
    \toprule
       \multirow{2}{*}{\textbf{Models}}  & \multicolumn{3}{c||}{\textbf{ATIS}}  & \multicolumn{3}{c}{\textbf{SNIPS}}\\
    \cline{2-7}
         & \textbf{Intent Acc} & \textbf{Slot F1} & \textbf{Overall Acc}  & \textbf{Intent Acc} & \textbf{Slot F1} & \textbf{Overall Acc} \\
    \midrule\midrule
        Attention BiRNN\cite{liu_attention-based_2016} & 91.1 & 94.2& 78.9  & 96.7 & 87.8& 74.1 \\ \midrule
        Slot-Gated Full Atten\cite{goo_slot-gated_2018} & 93.6 & 94.8& 82.2  & 97.0 & 88.8& 75.5 \\
        Graph LSTM\cite{zhang_graph_2020} & 97.2& 95.9& 87.5& 98.2& 95.3& 89.7 \\
        Co-Interactive transformer\cite{qin_co-interactive_2021} & 97.7& 95.9& 87.4 & 98.8& 95.9& 90.3 \\
        SyntacticTF(Independent) \cite{wang2021encoding} &     \textbf{98.1} & 95.9 & - & 98.7 & 96.5 & -  \\
        DGM \cite{ding2021focus} & 97.4 & \underline{96.1} & 87.8 & 98.2 & 85.2 & 88.4\\
        \midrule
        Joint BERT\cite{chen_bert_2019} & \underline{97.8}& 95.7& 88.2  & \underline{99.0}& 96.2& 91.6 \\
        Stack-Propgation+BERT\cite{qin_stack-propagation_2019} & 97.5 & \underline{96.1} & \underline{88.6} &  \underline{99.0}& \underline{97.0} & \textbf{92.9}\\
        \midrule
        \textbf{ESIS-BERT} & \textbf{98.1} & \textbf{98.2} & \textbf{90.1} & \textbf{99.1} & \textbf{97.8} & 92.1 \\
        \bottomrule
    \end{tabular}}
    \label{tab:ex-result}
\end{table}

\section{Experimental Analysis}
In this section, we conduct an ablation analysis and describe the effectiveness of major components of the ESIS-BERT.
Thus, we design experiments to illustrate the efficacy of the sub-words attention adapter, the intent attention adapter.
And, we analyze the effect of intent features for slot prediction.
Furthermore, we describe the analysis of model results and a case study on ATIS \cite{hemphill1990atis} datasets. 
In addition, we use the results of the visualization to show the process of learning the SAA.

\subsection{Effect of Sub-words attention adapter for Slot Filling}
\label{section-EFF-WTAtt}
We introduce the SAA to capture the sub-tokens contextual features for complex words in a sentence. 
We train the ESIS-BERT method only on the slot filling task, in order to evaluate the effectiveness of the SAA. 
The main ablation experimental design is as follows:

\begin{itemize}
\item \textbf{w/o the intent classification task.}
We train the only for the slot tagging task. 
No further joint optimization for this experiment, only the slot tagging task.
The slot decoder is not dependent on the intent information from IAA and only feeds the output states from the SAA or the BERT into the slot decoder.
\end{itemize}

The ESIS-BERT (for slot) row in Table \ref{tab:ablation} shows that the slot filling F1 has further improved while revealing that the sub-words attention adapter aids slot filling task. 
The slot filling task achieves significant results without intent classification task, there is an improvement of 0.5\% and 0.2\% over the ESIS-BERT results on the ATIS and SNIPS datasets, respectively.

\subsection{Ablation Analysis} 
\label{section-AA}
To analyze the effectiveness of the sub-words attention adapter, the intent attention adapter, and the intent feature for the slot in the ESIS-BERT model, we design three ablation experimental designs as follows:

\begin{itemize}

\item \textbf{w/o the sub-words attention adapter.} 
The SAA is removed, and for each complex token, the hidden state of the first sub-token is used directly to predict its slot label.

\item \textbf{w/o the intent attention adapter.} 
The IAA is removed, and for user intention, the hidden state of the token ([CLS]) is used directly to predict its intent label.

\item \textbf{w/o the intent feature for slot prediction.} 
Only feeds hidden states from SAA into slot decoder, the slot decoder is predicted without adding the intention information from IAA.
\end{itemize}

Table \ref{tab:ablation} shows the results of the ESIS-BERT method using the ATIS and SNIPS datasets after one component or much information were removed. 
It can be clearly seen that each layer and component of the model is useful in improving performance. 
Upon removing the IAA, the score of slot F1 and overall accuracy decreased for both datasets. 
Experimental results show that the IAA has a significant effect on the result of the ESIS-BERT. 
Upon removing the SAA, the performance on both datasets decreases dramatically. 
With the ATIS dataset, slot F1 score and overall accuracy decrease by 5.4\% and 8.8\% and decrease by 2.2\% and 2.1\% for the SNIPS dataset. 
The above results further illustrate the SAA's effectiveness, which can extract the sub-tokens contextual information in complex words to aid in the slot filling task.
Upon removing the intent feature, the accuracy scores of intention are almost the same as the ESIS-BERT results but the scores of slot F1 and overall accuracy decreased. 
These results show that intention features aid slot tagging task.

\begin{table}[t]
    \centering
    \caption{\textbf{Ablation analysis.} The first group is the result of Section \ref{section-EFF-WTAtt}. The second group is the result of Section 
    \ref{section-AA}. The third group is the result of the ESIS-BERT.}
    \resizebox{\textwidth}{!}{
    \begin{tabular}{l || ccc || ccc}
    \toprule
       \multirow{2}{*}{\textbf{Models}}  & \multicolumn{3}{c||}{\textbf{ATIS}}  & \multicolumn{3}{c}{\textbf{SNIPS}}\\
    \cline{2-7}
         & \textbf{Intent Acc} & \textbf{Slot F1} & \textbf{Overall Acc}  & \textbf{Intent Acc} & \textbf{Slot F1} & \textbf{Overall Acc} \\
    \midrule
    ESIS-BERT (for slot) & - & \textbf{98.7} & - & - & \textbf{98.0} & - \\
    \midrule
    w/o IAA & 97.5 & 97.1 & 88.0 & 98.6 & 97.0 & 90.4 \\
    w/o SAA & 97.7 & 92.8 & 81.3 & 98.2 & 95.6 & 90.0 \\
    w/o intent feature & 97.8 & 97.9 & 88.4 & 98.7 & 97.6 & 91.0 \\
    \midrule
        \textbf{ESIS-BERT} & \textbf{98.1} & \textbf{98.2} & \textbf{90.1} & \textbf{99.1} & \textbf{97.8} & \textbf{92.1} \\
        \bottomrule
    \end{tabular}}
    \label{tab:ablation}
\end{table}

\subsection{Model Result Analysis}
\begin{figure}[H]
    \centering
    \subfigure[Dfferent epochs with the ATIS dataset.]{
        \centering
        \includegraphics[width=0.8\textwidth]{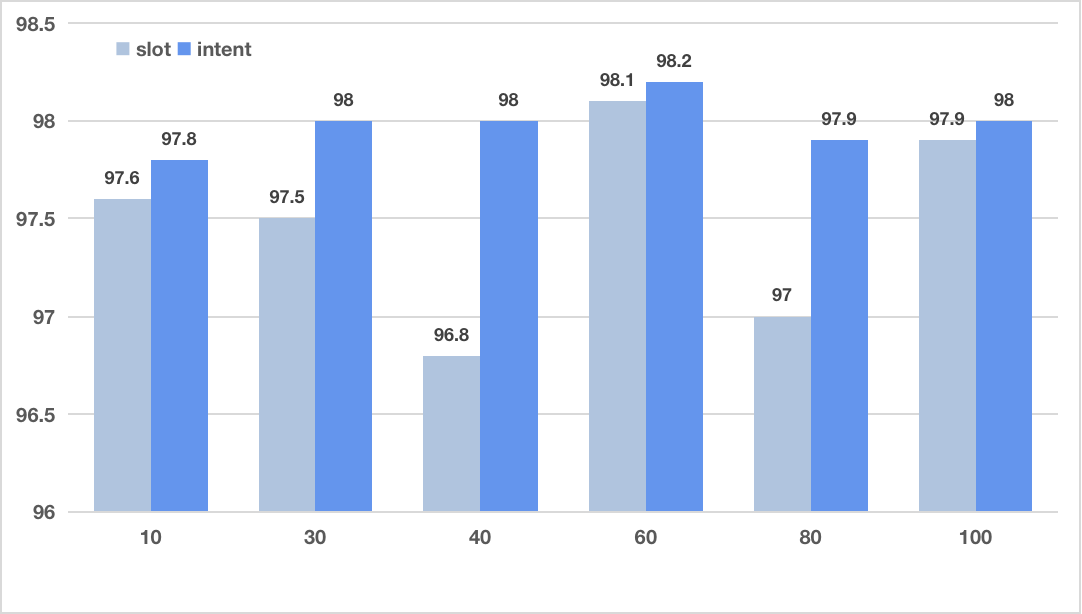}
    }
    \subfigure[Different epochs with the SNIPS dataset.]{
    \centering
    \includegraphics[width=0.8\textwidth]{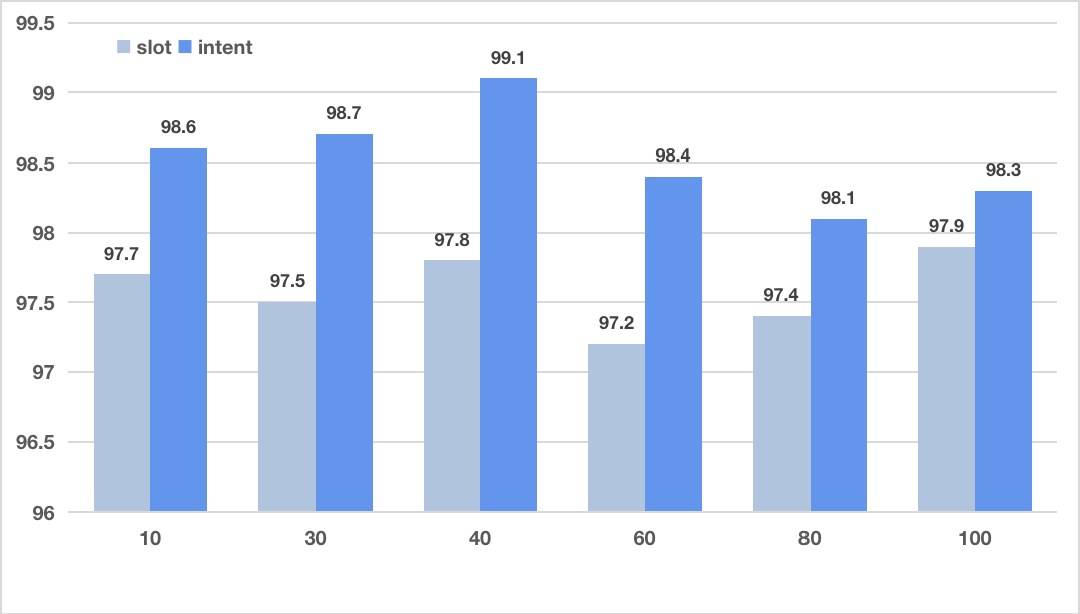}
    }
    \caption{\textbf{Analysis with different epochs and the (a) ATIS dataset, and the (b) SNIPS dataset.} The epochs are represented on the X-axis, and the scores of intent acc and slot F1 are represented on the Y-axis.}
    \label{fig:epochs}
\end{figure}

To select the best set of results obtained using the ATIS and SNIPS datasets, we analyze different epochs (10, 30, 40, 60, 80, and 100) for the ESIS-BERT. 
Figure \ref{fig:epochs} shows that with the ATIS dataset, the best intent acc is 98.1\% and the best slot F1 is 98.2\% by epoch 60. 
Figure \ref{fig:epochs} (b) shows that with the SNIPS datasets, best intent acc and slot F1 are 99.1\% and 97.8\%, respectively, by epoch 40.


\subsection{Case Study}
To compare the ESIS-BERT model with other BERT-based models, we select a case from ATIS, as shown in Table \ref{tab:case-study}. 
ESIE-BERT outperforms the joint BERT model \cite{chen_bert_2019} by capturing the sub-tokens contextual information of the SAA. 
In this case, the \textit{"june thirtieth"}  is incorrectly predicted by the joint BERT as an \textit{O}. 
However, the ESIE-BERT model correctly predicts the slot labels because “june thirtieth” is a complex word and the SAA worked well.

\begin{table}[htb]
\caption{\textbf{Cases in the ATIS dataset.} The slots of \emph{"june"} and \emph{"thirtieth"} are \emph{"B-depart\_date.month\_name"} and \emph{"B-depart\_date.day\_number"}. The green parts are correctly predicted by our ESIE-BERT method, and the red parts are the error predicted by the joint BERT model.}
    \centering
        \resizebox{\textwidth}{!}{
    \begin{tabular}{l||l}
    \hline
        Query &  what is the \textbf{earliest} flight from \textbf{memphis} to \textbf{cincinnati} on \textbf{june thirtieth}\\ 
        Wordpiece & 'what', 'is', 'the', 'earliest', 'flight', 'from', 'memphis', 'to', 'cincinnati', 'on', 'june', 'th', '\#\#ir', '\#\#tie', '\#\#th'\\ 
        \hline\hline
        \multicolumn{2}{l}{\textbf{Correctly}, predicted by the ESIE-BERT}\\ \hline
        Intent & atis\_flight \\
        Slots & O O O B-flight\_mod O O B-fromloc.city\_name O B-toloc.city\_name O \textcolor{green}{\textbf{B-depart\_date.month\_name B-depart\_date.day\_number}}\\
        \hline \hline
        \multicolumn{2}{l}{Predicted by the joint BERT \cite{chen_bert_2019}}\\ \hline
        Intent & atis\_flight \\
        Slots & O O O B-flight\_mod O O B-fromloc.city\_name O B-toloc.city\_name O \textcolor{red}{\textbf{O O}}\\
        \hline 
    \end{tabular} }
    
    \label{tab:case-study}
\end{table}

\subsection{Visualization}
\begin{figure}[H]
    \centering
\includegraphics[width=0.8\textwidth]{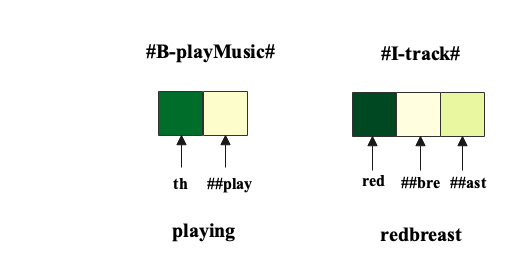}
    \caption{\textbf{Visualization.} Sub-tokens of \emph{"playing"} is \emph{["play", "\#\#ing"]} and the slot label is \emph{"I-day"}. Sub-tokens of \emph{"redbreast"} is \emph{["red", "\#\#bre", "\#\#ast"]} and the slot label is \emph{"I-track"}. The darker the color of each sub-token, the more features it feeds.}
    \label{fig:vis}
\end{figure}

We visualize the attention weights of the subwords in the SAA, in order to gain some insight into what the proposed method has learned, as shown in Figure \ref{fig:vis}.
From the complex words "playing" and "redbreast" and the slots "B-playMusic" and "I-track", it is evident that the attention weights of each complex word are focused on complete sub-tokens, which reveal complete sub-tokens information aid slot filling.
This indicates that the wordpiece tokenization attention mechanism helps the ESIS-BERT model construct complete sentence semantic information. 

\section{Conclusion and Future Work}
In this paper, we propose a novel joint method, named the ESIS-BERT model, for intention detection and slot tagging that explicitly models multiple sub-tokens’ features after wordpiece algorithm. 
The ESIS-BERT is introduced to solve two key issues of wordpiece tokenization on top of BERT: 
(1) How to predict slot tags without
knowing the number of sub-tokens in advance?
(2) How to capture sub-tokens contextual features of complex words? 
Experiments show that the ESIS-BERT outperforms the SOTA methods, demonstrating the efficacy of our proposed SAA and IAA for the NLU tasks.
Compared with previous SOTA methods, our proposed ESIE-BERT also indicates significant performance improvements in intent classification accuracy, slot filling F1, and sentence-level accuracy on the ATIS and SNIPS datasets.
Finally, compared with other models, the ESIS-BERT is more useful for slot tagging, due to the relationships between complete sub-tokens being stronger and contextual information being easily modeled.
This work will help guide pre-model design for more NLU problems.
In the future, our work will attempt to perform more sequence labeling tasks by using the ESIS-BERT model and further model sub-tokens information.

\section*{Acknowledgement}
The research is supported by the National Natural Science Foundation of China under Grant Nos. 61906159, 62176014, 71873108, 62072379, Natural Science Foundation of Sichuan Province under Grant No. 2023NSFSC0032, 2023NSFSC0114, and Guanghua Talent Project of Southwestern University of Finance and Economics, Financial Innovation Center, SWUFE (Project NO.FIC2022C0008) and ``Double-First Class" International Innovation Project (SYL22GJCX07).

\bibliography{main.bib}

\begin{thebibliography}{10}
\expandafter\ifx\csname url\endcsname\relax
  \def\url#1{\texttt{#1}}\fi
\expandafter\ifx\csname urlprefix\endcsname\relax\def\urlprefix{URL }\fi
\expandafter\ifx\csname href\endcsname\relax
  \def\href#1#2{#2} \def\path#1{#1}\fi

\bibitem{zhao2022Stock}
Y.~Zhao, H.~Du, Y.~Liu, S.~Wei, X.~Chen, F.~Zhuang, Q.~Li, G.~Kou, Stock
  movement prediction based on bi-typed hybrid-relational market knowledge
  graph via dual attention networks, IEEE Transactions on Knowledge \& Data
  Engineering~(01) (2022) 1--12.

\bibitem{zhao2022learning}
Y.~Zhao, S.~Wei, H.~Du, X.~Chen, Q.~Li, F.~Zhuang, J.~Liu, G.~Kou, Learning
  bi-typed multi-relational heterogeneous graph via dual hierarchical attention
  networks, IEEE Transactions on Knowledge \& Data Engineering~(01) (2022)
  1--12.

\bibitem{tan2022finhgnn}
J.~Tan, Q.~Li, J.~Wang, J.~Chen, Finhgnn: A conditional heterogeneous graph
  learning to address relational attributes for stock predictions, Information
  Sciences.

\bibitem{huang2022asset}
J.~Huang, R.~Xing, Q.~Li, Asset pricing via deep graph learning to incorporate
  heterogeneous predictors, International Journal of Intelligent Systems
  37~(11) (2022) 8462--8489.

\bibitem{tur2011intent}
G.~Tur, L.~Deng, Intent determination and spoken utterance classification,
  Spoken language understanding: systems for extracting semantic information
  from speech (2011) 93--118.

\bibitem{tur2011spoken}
G.~Tur, R.~De~Mori, Spoken language understanding: Systems for extracting
  semantic information from speech, John Wiley \& Sons, 2011.

\bibitem{hemphill1990atis}
C.~T. Hemphill, J.~J. Godfrey, G.~R. Doddington, The atis spoken language
  systems pilot corpus, in: Speech and Natural Language: Proceedings of a
  Workshop Held at Hidden Valley, Pennsylvania, June 24-27, 1990, 1990.

\bibitem{ramanand2010wishful}
J.~Ramanand, K.~Bhavsar, N.~Pedanekar, Wishful thinking-finding suggestions
  and’buy’wishes from product reviews, in: Proceedings of the NAACL HLT
  2010 workshop on computational approaches to analysis and generation of
  emotion in text, 2010, pp. 54--61.

\bibitem{schuurmans2019intent}
J.~Schuurmans, F.~Frasincar, Intent classification for dialogue utterances,
  IEEE Intelligent Systems 35~(1) (2019) 82--88.

\bibitem{goo_slot-gated_2018}
C.-W. Goo, G.~Gao, Y.-K. Hsu, C.-L. Huo, T.-C. Chen, K.-W. Hsu, Y.-N. Chen,
  Slot-gated modeling for joint slot filling and intent prediction, in:
  Proceedings of the 2018 Conference of the North American Chapter of the
  Association for Computational Linguistics: Human Language Technologies,
  Volume 2 (Short Papers), 2018, pp. 753--757.

\bibitem{guo2014joint}
D.~Guo, G.~Tur, W.-t. Yih, G.~Zweig, Joint semantic utterance classification
  and slot filling with recursive neural networks, in: 2014 IEEE Spoken
  Language Technology Workshop (SLT), IEEE, 2014, pp. 554--559.

\bibitem{hakkani2016multi}
D.~Hakkani-T{\"u}r, G.~T{\"u}r, A.~Celikyilmaz, Y.-N. Chen, J.~Gao, L.~Deng,
  Y.-Y. Wang, Multi-domain joint semantic frame parsing using bi-directional
  rnn-lstm., in: Interspeech, 2016, pp. 715--719.

\bibitem{devlin2018bert}
J.~Devlin, M.-W. Chang, K.~Lee, K.~Toutanova, Bert: Pre-training of deep
  bidirectional transformers for language understanding, arXiv preprint
  arXiv:1810.04805.

\bibitem{chen_bert_2019}
Q.~Chen, Z.~Zhuo, W.~Wang, Bert for joint intent classification and slot
  filling, arXiv preprint arXiv:1902.10909.

\bibitem{feng2022pretraining}
Z.~Feng, D.~Tang, C.~Zhou, J.~Liao, S.~Wu, X.~Feng, B.~Qin, Y.~Cao, S.~Shi,
  Pretraining without wordpieces: learning over a vocabulary of millions of
  words, arXiv preprint arXiv:2202.12142.

\bibitem{yang_aise_2021}
P.~Yang, D.~Ji, C.~Ai, B.~Li, Aise: Attending to intent and slots explicitly
  for better spoken language understanding, Knowledge-Based Systems 211 (2021)
  106537.

\bibitem{coucke2018snips}
A.~Coucke, A.~Saade, A.~Ball, T.~Bluche, A.~Caulier, D.~Leroy, C.~Doumouro,
  T.~Gisselbrecht, F.~Caltagirone, T.~Lavril, et~al., Snips voice platform: an
  embedded spoken language understanding system for private-by-design voice
  interfaces, arXiv preprint arXiv:1805.10190.

\bibitem{abdalla2022integration}
H.~I. Abdalla, A.~A. Amer, On the integration of similarity measures with
  machine learning models to enhance text classification performance,
  Information Sciences 614 (2022) 263--288.

\bibitem{cui2022self}
H.~Cui, G.~Wang, Y.~Li, R.~E. Welsch, Self-training method based on gcn for
  semi-supervised short text classification, Information Sciences 611 (2022)
  18--29.

\bibitem{wu2017efficient}
Z.~Wu, H.~Zhu, G.~Li, Z.~Cui, H.~Huang, J.~Li, E.~Chen, G.~Xu, An efficient
  wikipedia semantic matching approach to text document classification,
  Information Sciences 393 (2017) 15--28.

\bibitem{kim_convolutional_2014}
Y.~Kim, \href{https://aclanthology.org/D14-1181}{Convolutional neural networks
  for sentence classification}, in: Proceedings of the 2014 Conference on
  Empirical Methods in Natural Language Processing ({EMNLP}), Association for
  Computational Linguistics, Doha, Qatar, 2014, pp. 1746--1751.
\newblock \href {http://dx.doi.org/10.3115/v1/D14-1181}
  {\path{doi:10.3115/v1/D14-1181}}.
\newline\urlprefix\url{https://aclanthology.org/D14-1181}

\bibitem{zhang_character-level_2015}
X.~Zhang, J.~Zhao, Y.~LeCun, Character-level convolutional networks for text
  classification, Advances in neural information processing systems 28.

\bibitem{ravuri_recurrent_nodate}
S.~Ravuri, A.~Stolcke, Recurrent neural network and lstm models for lexical
  utterance classification, in: Sixteenth Annual Conference of the
  International Speech Communication Association, 2015.

\bibitem{zhao_attention-based_2016}
Z.~Zhao, Y.~Wu, Attention-based convolutional neural networks for sentence
  classification., in: Interspeech, Vol.~8, 2016, pp. 705--709.

\bibitem{yang_hierarchical_2016}
Z.~Yang, D.~Yang, C.~Dyer, X.~He, A.~Smola, E.~Hovy, Hierarchical attention
  networks for document classification, in: Proceedings of the 2016 conference
  of the North American chapter of the association for computational
  linguistics: human language technologies, 2016, pp. 1480--1489.

\bibitem{zhang2023strengthened}
R.~Zhang, S.~Luo, L.~Pan, Y.~Ma, Z.~Wu, Strengthened multiple correlation for
  multi-label few-shot intent detection, Neurocomputing 523 (2023) 191--198.

\bibitem{mccallum2000maximum}
A.~McCallum, D.~Freitag, F.~C. Pereira, Maximum entropy markov models for
  information extraction and segmentation., in: Icml, Vol.~17, 2000, pp.
  591--598.

\bibitem{raymond2007generative}
C.~Raymond, G.~Riccardi, Generative and discriminative algorithms for spoken
  language understanding, in: Interspeech 2007-8th Annual Conference of the
  International Speech Communication Association, 2007.

\bibitem{vu_sequential_2016}
N.~T. Vu, Sequential convolutional neural networks for slot filling in spoken
  language understanding, arXiv preprint arXiv:1606.07783.

\bibitem{yao_spoken_2014}
K.~Yao, B.~Peng, Y.~Zhang, D.~Yu, G.~Zweig, Y.~Shi, Spoken language
  understanding using long short-term memory neural networks, in: 2014 IEEE
  Spoken Language Technology Workshop (SLT), IEEE, 2014, pp. 189--194.

\bibitem{peng2015recurrent}
B.~Peng, K.~Yao, L.~Jing, K.-F. Wong, Recurrent neural networks with external
  memory for spoken language understanding, in: Natural Language Processing and
  Chinese Computing, Springer, 2015, pp. 25--35.

\bibitem{kurata_leveraging_2016}
G.~Kurata, B.~Xiang, B.~Zhou, M.~Yu, Leveraging sentence-level information with
  encoder lstm for semantic slot filling, arXiv preprint arXiv:1601.01530.

\bibitem{zhao_improving_2018}
L.~Zhao, Z.~Feng, Improving slot filling in spoken language understanding with
  joint pointer and attention, in: Proceedings of the 56th Annual Meeting of
  the Association for Computational Linguistics (Volume 2: Short Papers), 2018,
  pp. 426--431.

\bibitem{he2021context}
K.~He, Y.~Yan, W.~Xu, From context-aware to knowledge-aware: boosting oov
  tokens recognition in slot tagging with background knowledge, Neurocomputing
  445 (2021) 267--275.

\bibitem{liu_attention-based_2016}
B.~Liu, I.~Lane, Attention-based recurrent neural network models for joint
  intent detection and slot filling, arXiv preprint arXiv:1609.01454.

\bibitem{zhang_graph_2020}
L.~Zhang, D.~Ma, X.~Zhang, X.~Yan, H.~Wang, Graph lstm with context-gated
  mechanism for spoken language understanding, in: Proceedings of the AAAI
  Conference on Artificial Intelligence, Vol.~34, 2020, pp. 9539--9546.

\bibitem{qin_co-interactive_2021}
L.~Qin, T.~Liu, W.~Che, B.~Kang, S.~Zhao, T.~Liu, A co-interactive transformer
  for joint slot filling and intent detection, in: ICASSP 2021-2021 IEEE
  International Conference on Acoustics, Speech and Signal Processing (ICASSP),
  IEEE, 2021, pp. 8193--8197.

\bibitem{wang2021encoding}
J.~Wang, K.~Wei, M.~Radfar, W.~Zhang, C.~Chung, Encoding syntactic knowledge in
  transformer encoder for intent detection and slot filling, in: Proceedings of
  the AAAI Conference on Artificial Intelligence, Vol.~35, 2021, pp.
  13943--13951.

\bibitem{tang2020end}
H.~Tang, D.~Ji, Q.~Zhou, End-to-end masked graph-based crf for joint slot
  filling and intent detection, Neurocomputing 413 (2020) 348--359.

\bibitem{ding2021focus}
Z.~Ding, Z.~Yang, H.~Lin, J.~Wang, Focus on interaction: A novel dynamic graph
  model for joint multiple intent detection and slot filling., in: IJCAI, 2021,
  pp. 3801--3807.

\bibitem{DBLP:journals/corr/abs-1802-05365}
M.~E. Peters, M.~Neumann, M.~Iyyer, M.~Gardner, C.~Clark, K.~Lee,
  L.~Zettlemoyer, \href{http://arxiv.org/abs/1802.05365}{Deep contextualized
  word representations}, CoRR abs/1802.05365.
\newblock \href {http://arxiv.org/abs/1802.05365} {\path{arXiv:1802.05365}}.
\newline\urlprefix\url{http://arxiv.org/abs/1802.05365}

\bibitem{radford2018improving}
A.~Radford, K.~Narasimhan, T.~Salimans, I.~Sutskever, et~al., Improving
  language understanding by generative pre-training.

\bibitem{qin_stack-propagation_2019}
L.~Qin, W.~Che, Y.~Li, H.~Wen, T.~Liu, A stack-propagation framework with
  token-level intent detection for spoken language understanding, arXiv
  preprint arXiv:1909.02188.

\bibitem{lample2019cross}
G.~Lample, A.~Conneau, Cross-lingual language model pretraining, arXiv preprint
  arXiv:1901.07291.

\bibitem{joshi2020spanbert}
M.~Joshi, D.~Chen, Y.~Liu, D.~S. Weld, L.~Zettlemoyer, O.~Levy, Spanbert:
  Improving pre-training by representing and predicting spans, Transactions of
  the Association for Computational Linguistics 8 (2020) 64--77.

\bibitem{yang2019xlnet}
Z.~Yang, Z.~Dai, Y.~Yang, J.~Carbonell, R.~R. Salakhutdinov, Q.~V. Le, Xlnet:
  Generalized autoregressive pretraining for language understanding, Advances
  in neural information processing systems 32.

\bibitem{lan2019albert}
Z.~Lan, M.~Chen, S.~Goodman, K.~Gimpel, P.~Sharma, R.~Soricut, Albert: A lite
  bert for self-supervised learning of language representations, arXiv preprint
  arXiv:1909.11942.

\bibitem{kingma2014adam}
D.~P. Kingma, J.~Ba, Adam: A method for stochastic optimization, arXiv preprint
  arXiv:1412.6980.

\bibitem{qin_survey_2021}
L.~Qin, T.~Xie, W.~Che, T.~Liu, A survey on spoken language understanding:
  Recent advances and new frontiers, arXiv preprint arXiv:2103.03095.

\end{thebibliography}

\end{document}